\documentclass[letterpaper, 10 pt, conference]{ieeeconf} 

\IEEEoverridecommandlockouts                              
\usepackage{graphicx}   
\usepackage{amsmath} 
\usepackage{amssymb}  
\usepackage{kantlipsum}
\usepackage{mathtools, cuted}
\usepackage{units}
\usepackage{comment}
\usepackage{color}  

\title{\LARGE \bf Design and Control of a Recovery System for Legged Robots}

\author{Kevin Green, Nils Smit-Anseeuw, Rodney Gleason and C. David Remy, \emph{Member, IEEE}
\thanks{The authors are with the Robotics and Motion Laboratory (RAMlab), Department of Mechanical Engineering, University of Michigan, Ann Arbor, MI {\tt \small (kevingre@umich.edu, nilssmit@umich.edu, gleasonr@umich.edu , cdremy@umich.edu)}.  This material is based upon work supported by the National Science Foundation under Grant No. 1453346. Any opinion, findings, and conclusions or recommendations expressed in this material are those of the authors(s) and do not necessarily reflect the views of the National Science Foundation.}}

\begin{document}


\maketitle
\thispagestyle{empty}
\pagestyle{empty}
\begin{abstract}
This paper describes the design and control of a support and recovery system for use with planar legged robots.
The system operates in three modes. 
First, it can be operated in a fully transparent mode where no forces are applied to the robot.
In this mode, the system follows the robot closely to be able to quickly catch the robot if needed.
Second, it can provide a vertical supportive force to assist a robot during operation.
Third, it can catch the robot and pull it away from the ground after a failure to avoid falls and the associated damages.
In this mode, the system automatically resets the robot after a trial allowing for multiple consecutive trials to be run without manual intervention.
The supportive forces are applied to the robot through an actuated cable and pulley system that uses series elastic actuation with a unidirectional spring to enable truly transparent operation.
The nonlinear nature of this system necessitates careful design of controllers to ensure predictable, safe behaviors.
In this paper we introduce the mechatronic design of the recovery system, develop suitable controllers, and evaluate the system's performance on the bipedal robot RAM\emph{one}. 
\end{abstract}

\section{Introduction}
\label{sec:introduction}

Traditionally, falling is aggressively avoided in legged robotics research \cite{Karssen:2013}.
Falls constitute failure, resulting in the end of the test, and possible damage to the robot.
To continue, the roboticist must reset the robot, evaluate any damage, possibly recalibrate sensors, and restart the task.
This is a slow, cumbersome and risky procedure.
If a component of the robot breaks, it can delay testing days, weeks or even months.

The state of the art approach to catching falling planar robots is a hard stop. 
This approach has been used in boom systems by Hurst \cite{ hurst2007design}, Grizzle \cite{sreenath2011compliant}, Zeglin \cite{zeglin1991uniroo} and Geyer \cite{martin2015robust}.
This method is mechanically simple and cheap. 
There are many different possible implementations but they all support the robot if it falls lower than a set height.

This approach has several shortcomings. 
First, it is not able to automatically reset the robot after failure occurs.
If the robot falls often or if many trials are necessary, this can slow the trial procedure.
In particular, when online learning is implemented, many trials are needed and failures are common \cite{Kalakrishnan:2010a}.
Second, the failure condition of a minimum body height is overly simplistic.
The viability of a state is more complex than simply the current height \cite{Cnops2015}.
This simple failure condition can wind up halting the robot while it is still recoverable.
Third, this system does not have the ability to provide gentle assistance to the robot.
The system is either holding the robot or is not, with no middle ground.  

When children first learn to walk they are shadowed by a parent.
The parent will be close behind the child ready to catch the child if it is necessary.
Early in the learning process the parent may even help hold part of the child's weight.
This enables the child to experiment and learn how to move their legs without risking hurting themselves.
As the child learns to walk, the parent intervenes less and less until eventually they are not necessary.
We seek to incorporate similar assistance to legged robots.

\begin{figure} 
    \includegraphics[width=\columnwidth]{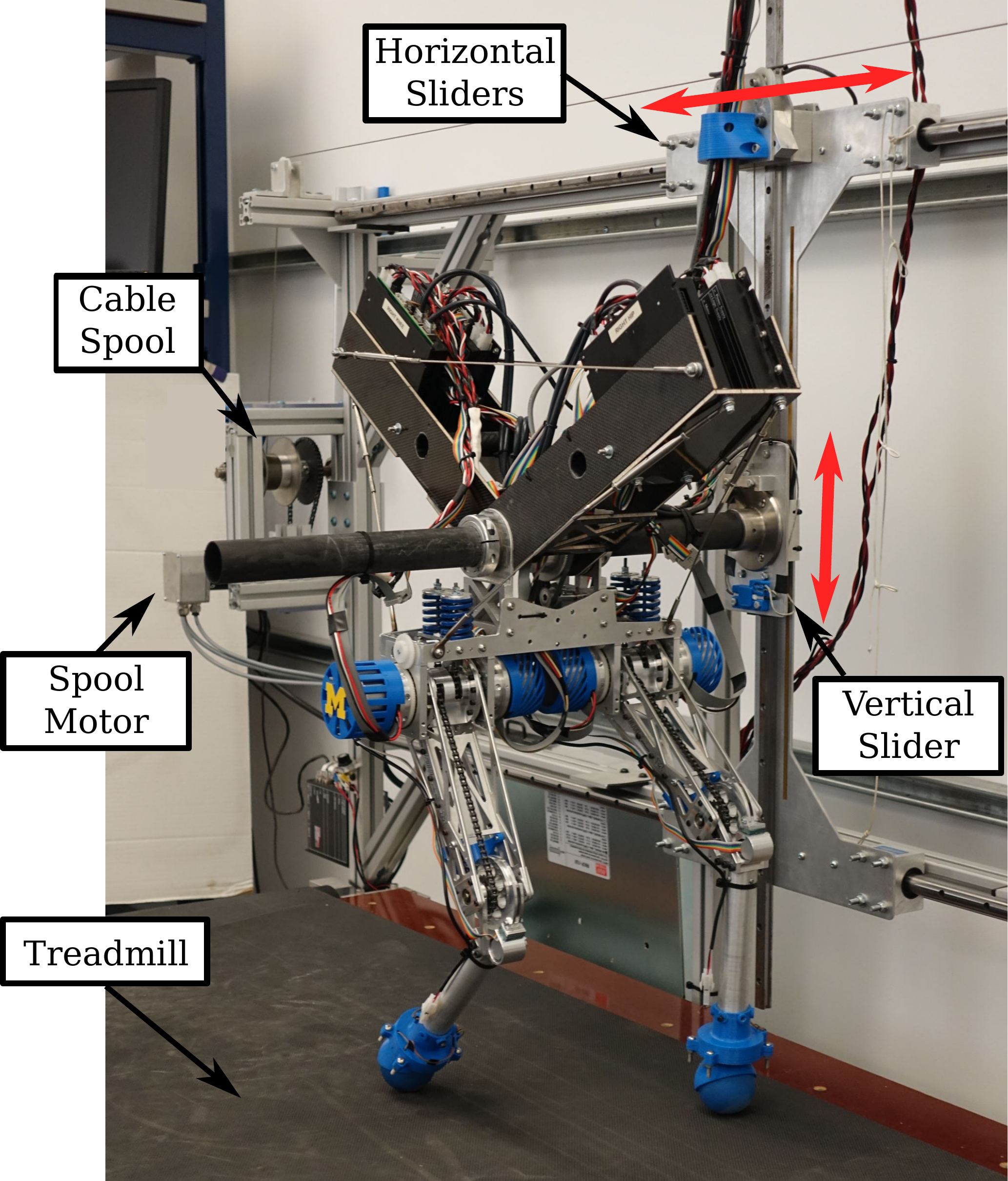}
    \caption{The Biped RAM\emph{one} and the planar support system. The pair of sliders enables the robot to move exclusively in the plane. The vertical slider is actuated by the powered winch through a pulley system.}
    \label{fig:ramone}
\end{figure}

To do this, we constructed a system to replicate the parent's helping hand for planar robots.
During operation it can have a minimal effect on the robot's dynamics.
In order to assist the robot's walking, it can also provide assistance in the form of a controlled vertical force.
When failure is detected, it can intervene immediately and reset the robot for the next experiment.

We begin this paper with a description of the mechatronic planarizer system (Section \ref{sec:MechatronicImplementation}). 
We then present a model of the system (Section \ref{sec:Model}).
In Section \ref{sec:ControllerDesign} we present a dynamically transparent controller, a programmable force controller, a method of failure detection, and a failure intervention controller.
Next, we evaluate the performance of each of the controllers (Section \ref{sec:results}).
Finally, we discuss our system's performance, our future use of the system and a concept for a boom implementation (Section \ref{sec:discussionandconclusion}).
\section{Mechatronic Implementation}
\label{sec:MechatronicImplementation}

The system that we designed actuates a linear bearing planarizer (Fig.~\ref{fig:ramone}), built
to keep our walking robots constrained to the plane.
To allow for continuous locomotion, the ground plane consists of a velocity controlled treadmill.
In this paper, the robot mounted to the fixture is the planar biped RAM\emph{one}.
RAM\emph{one} is a 5 link biped based on the ScarlETH leg \cite{hutter2011scarleth}.

The treadmill is powered by a \unit[2.38]{kW} 3-phase AC Motor.
The treadmill motor is controlled by a Leeson SM2 VFD (variable frequency drive) (model 174614.00).
The control signal for the VFD is supplied by an EtherCAT Beckhoff terminal block. 
The treadmill's transmission is a 5:1 speed reducing gearbox followed by a 2:1 timing belt speed reduction.
This timing belt drives a \unit[0.1016]{m} diameter urethane roller.
The roller is instrumented with a \unit[1000]{PPR} optical encoder (Broadcom Limited HEDM-5500\#B06) which enables feedback control.
The treadmill uses proportional feedback and open loop feedforward to maintain a constant desired velocity.
The treadmill is able to operate at a velocity between 0 and \unitfrac[1.91]{m}{sec}.

The robot is attached to the planarizer by a carbon fiber tube.
This tube is rigidly attached to a pair of linear bearings (Thomson 411N15A0) that can slide vertically along a single rail.
This rail is mounted to an aluminum cart that is supported at the top and bottom by four linear bearings (Thomson SSEPBOM20WW) which slide horizontally  along two rails.
In this arrangement the carbon fiber tube is free to move \unit[0.9]{m} vertically and \unit[1.8]{m} horizontally.
Both directions of travel are instrumented with \unit[10]{ $\mu$m} resolution Renishaw linear encoders (RGH41T).

\begin{figure} 
    \includegraphics[width=\columnwidth]{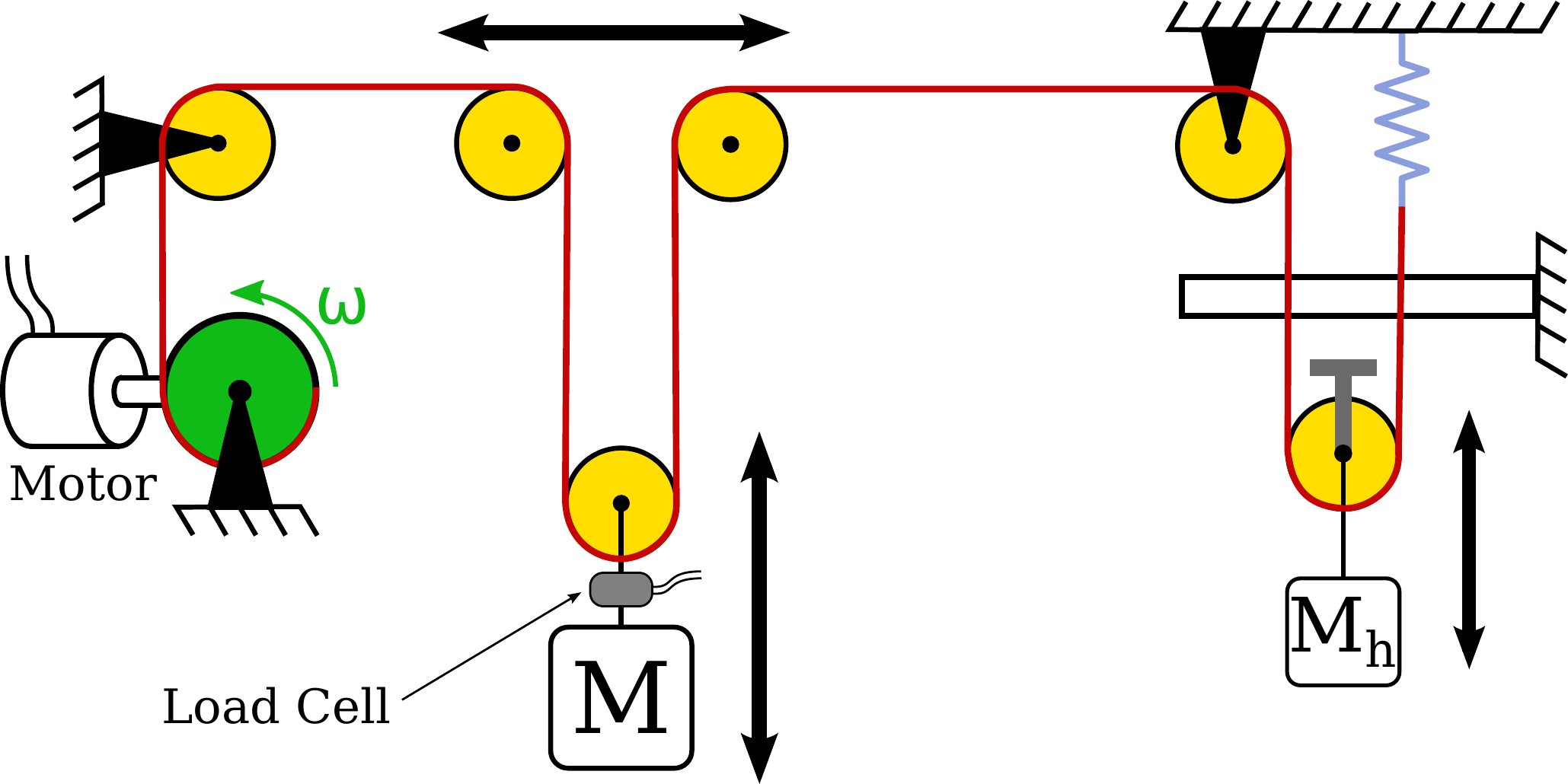}
    \caption{A diagram of the pulley system. The spool on the left is controlled to change the length of cable in the system. The robot is represented by $M$ and the hanging mass is represented by $M_h$. When the mass is lifted high enough it contacts a unidirectional compression spring.}
    \label{fig:sysDiagram}
\end{figure}

The active planarizer system only actuates the vertical slider, while allowing the cart to move freely horizontally.
This was accomplished through the use of a pulley system (Fig.~\ref{fig:sysDiagram}).
At the top of the cart there are a pair of pulleys that redirect the cable down to a moving pulley on the robot.
When the motor retracts the cable it will lift the robot vertically.
This pulley arrangement offers no resistance to horizontal movement with the exception of pulley friction and pulley inertia.
Additionally this arrangement allows the heavy motor to be placed off of the cart. 
On the far right hand side of the diagram, the end of the cable is fixed to a unidirectional extension spring.
This functions as a series elastic element between the motor and the robot, thereby improving force control performance \cite{Pratt:1995}.
As a cable driven system can only create pulling forces, the spring must only act in one direction.
Moreover, to prevent the cable from losing tension and jumping a pulley, a hanging mass is attached to the cable through a moving pulley. 
When the cable is loaded, this mass runs against a hard stop and engages the spring.
When the cable is losing tension, the mass slides downwards to take up the slack in the cable.
This system may appear overly complicated, but it adds minimal inertia to the robot, enables dynamically transparent tracking, and allows smooth force control.

\begin{figure} 
    \includegraphics[width=\columnwidth]{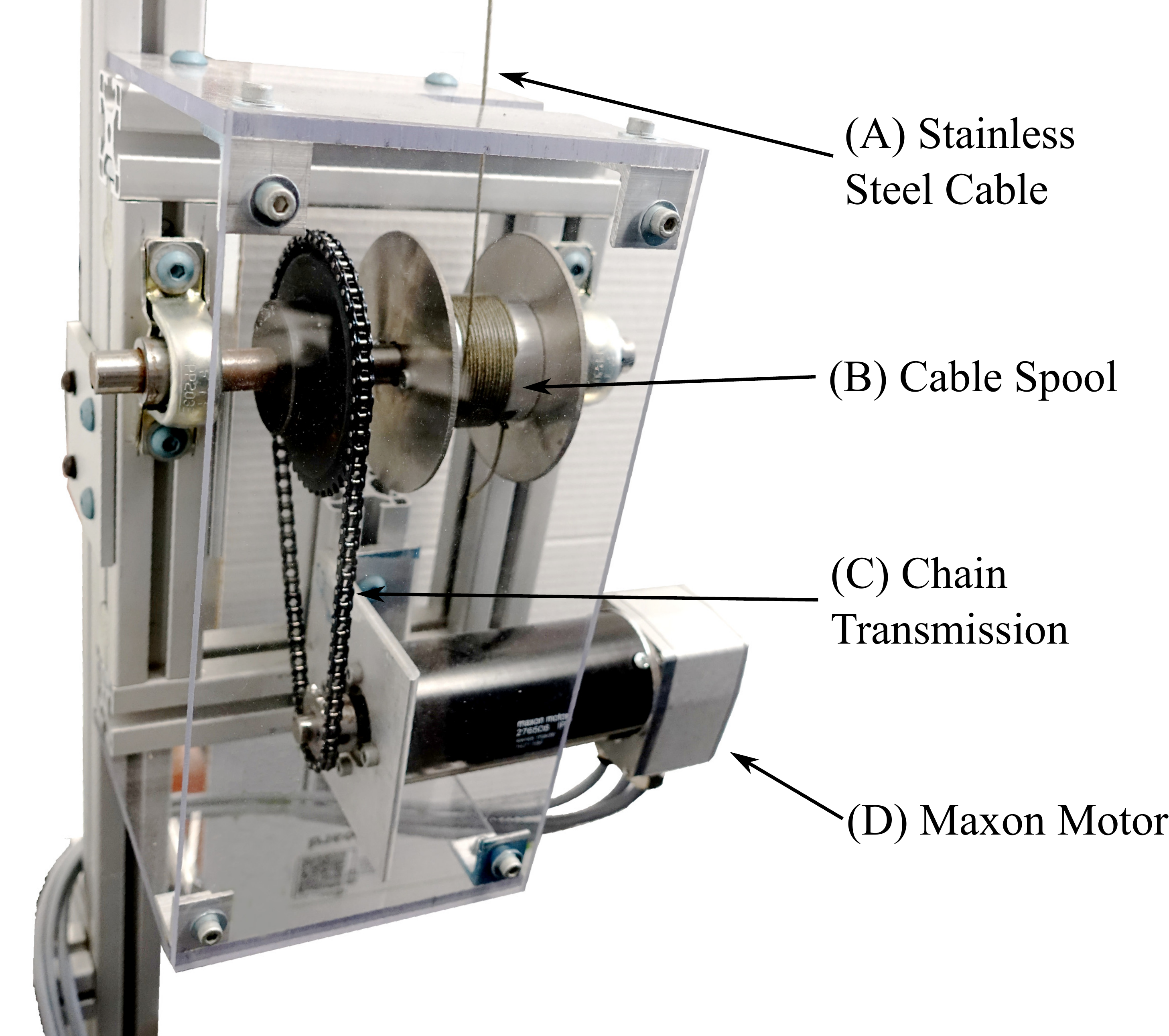}
    \caption{The motor and cable spool assembly. The spool winds or unwinds cable to lift, lower or apply force to the robot.}
    \label{fig:spoolPhoto}
\end{figure}

The cable (A) is nylon coated stainless steel wire with a diameter of \unit[1.59]{mm}.
The pulleys (F) are \unit[27.0]{mm} groove diameter nylon pulleys with integrated bearings.
This allows nearly frictionless operation of the pulley system.
The total length of cable in the system is controlled by a \unit[50.8]{mm} diameter powered spool (B fig. \ref{fig:spoolPhoto}) which  is driven by Maxon EC 60 brushless DC motor (D) through a 3:1 chain drive (C).
This motor is controlled by a Maxon EPOS3 digital positioning controller, which can perform on-board velocity control of the motor.
The robot is attached to a moving pulley through an Omega load cell (LC201-100) to measure the force that is applied to the robot by our system.
The load cell's measurements are read directly by a Beckhoff analog voltage input terminal.
This motor and spool system can apply up to \unit[500]{N} of force onto the robot with a maximum speed of \unitfrac[2.4]{m}{sec}.

\begin{figure} 
    \includegraphics[width=3 in]{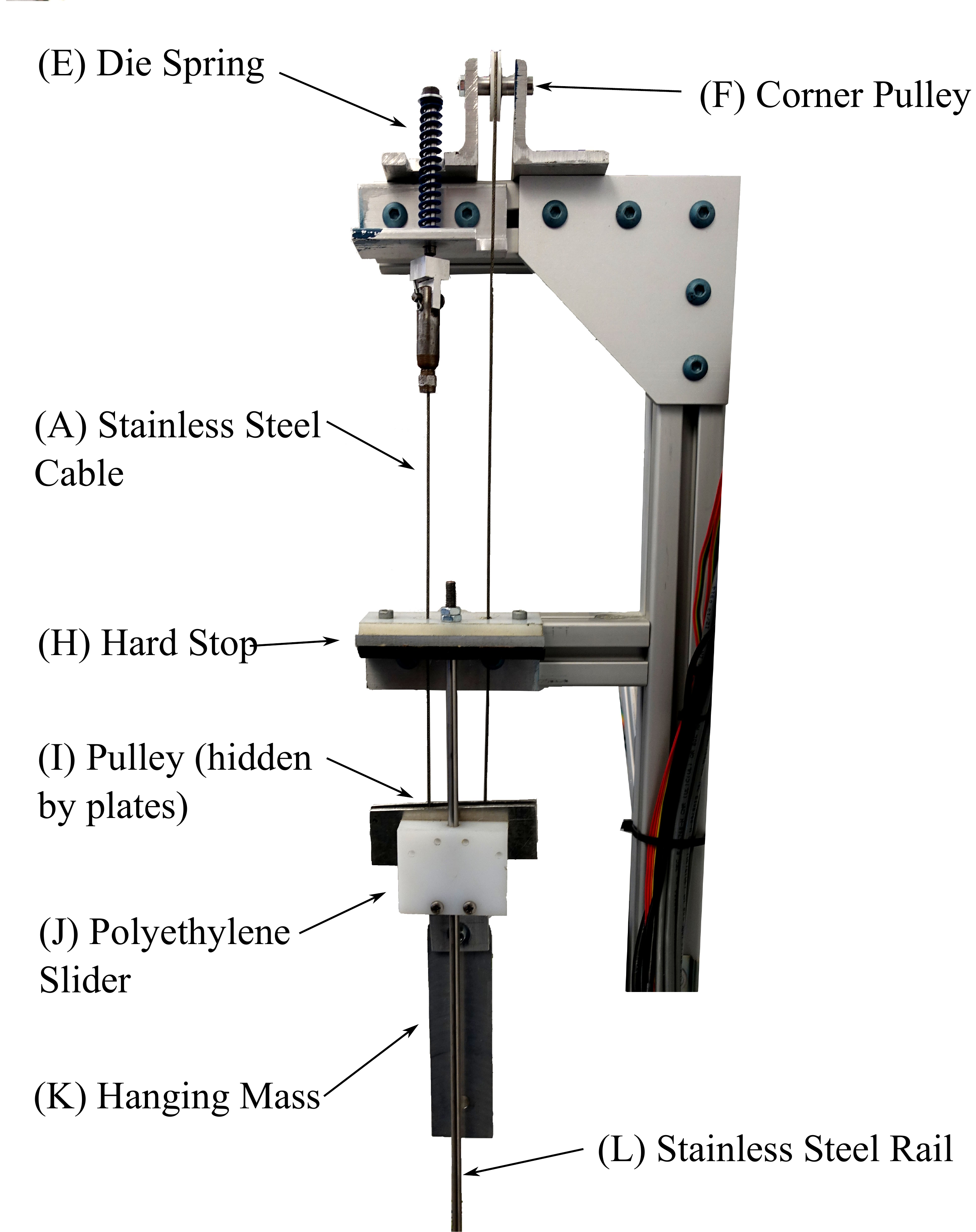}
    \caption{The hanging mass and spring assembly. The mass slides up and down on the stainless steel rail. The moving pulley is not visible because it is surrounded by two aluminum plates which contact the hard stop.}
    \label{fig:hangingMassPicture}
\end{figure}

The unidirectional spring is a die spring with a spring constant of \unitfrac[5250]{N}{m} (E fig. \ref{fig:hangingMassPicture}).
A rod runs through the center of the spring and is fastened to the end of the cable (A).
When the cable is loaded, the spring is compressed. 

The hanging mass (K) has a total mass of \unit[0.45]{kg}.
This mass was chosen to be large enough to ensure proper cable tension, yet small enough to minimize force disturbance on the robot.
The mass is constrained to move vertically on a \unit[6.35]{mm} diameter stainless steel rail (L).
The slider that runs along this rail is a custom slider made from UHMW Polyethylene (J).
A simple slider was chosen because the load on this bearing should be very small.
As long as the robot doesn't accelerate up faster than gravity, tension is maintained in the cable.
It is possible that acceleration greater than gravity is necessary, especially in running and hopping gaits.
If this is the case, a constant force spring can be used to replace the hanging mass.

The robot, planarizer and treadmill system is controlled by an EtherCAT (Ethernet for Control Automation Technology) fieldbus system. The DC brushless motors are controlled with Maxon EPOS3 motor controllers. All other inputs and output are routed through Beckhoff terminal blocks. The EtherCAT system is controlled by a desktop PC running MathWorks xPC Target.

\section{System Model}
\label{sec:Model}
\begin{figure} 
    \includegraphics[width=\columnwidth]{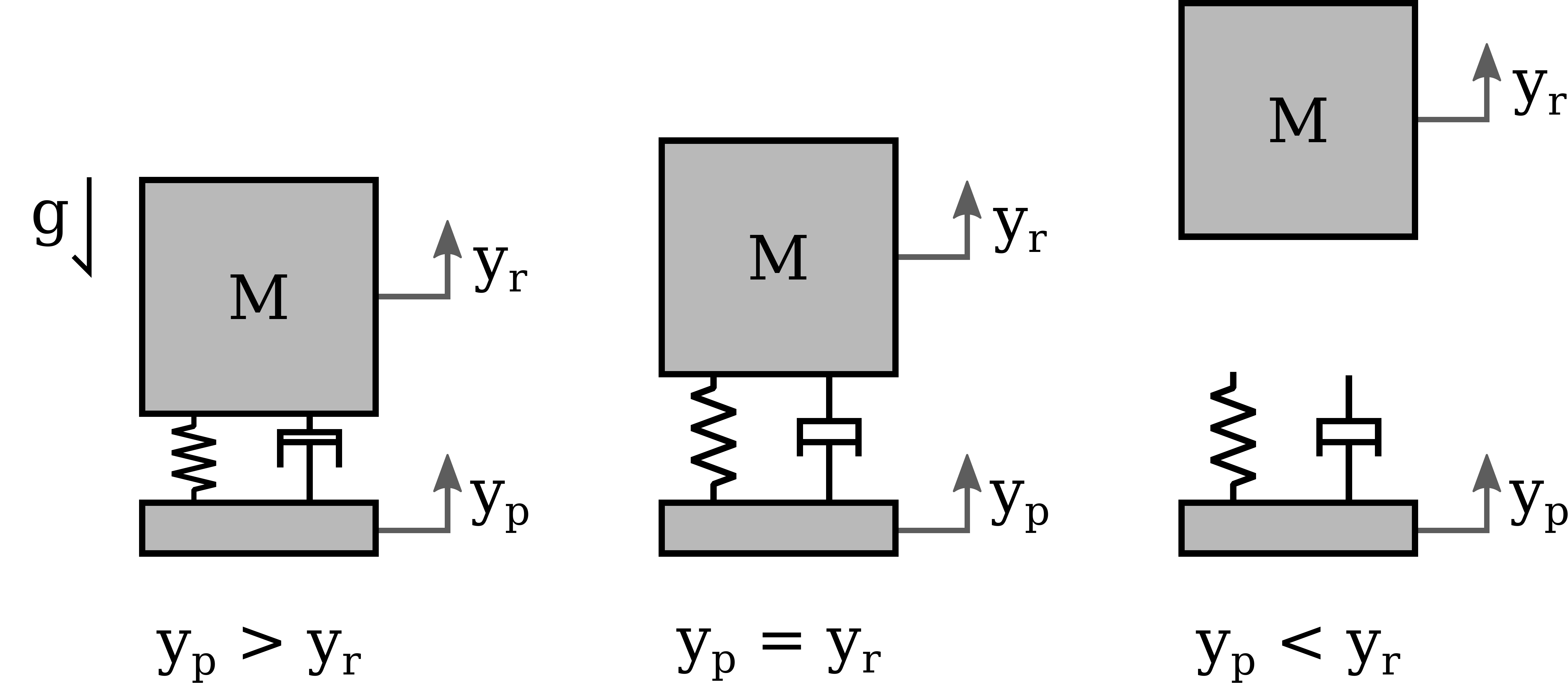}
    \caption{The simplified model of the planarizer system. The robot is allowed to lift off of the spring, which corresponds to the hanging mass dropping below the hard stop. This model does not account for effects resulting from the acceleration of the hanging mass. 
        }
    \label{fig:simpleSys}
\end{figure}

In order to better understand the underlying dynamics of the planarizer, a simple model is defined.
The model is shown in fig. \ref{fig:simpleSys} and the equation of motion is shown in eq. \ref{eq:EOM}.
The input $y_p$ is the planarizer motor position.
By multiplying by the transmission ratio between the robot and the motor, the motor position is moved into the robot's coordinate system. 
In this model, the spring is unidirectional just as the true spring is. 
\begin{equation}
    M \ddot{y}_r = F_{sp} + F_{ext} - F_g
    \label{eq:EOM}
\end{equation}
Here $M$ is the mass of the robot, $F_g$ is a modified gravitational force, $F_{ext}$ is the ground reaction forces on the robot's feet, and  $F_{sp}$ represents the elastic and damping forces from the spring.

The modified gravitational force is the resulting force on the robot from gravity combined with the weight of the hanging mass (eq. \ref{eq:FGrav}). The hanging mass has mass of $M_h$.
\begin{equation}
F_g = g(M - M_h)
    \label{eq:FGrav}
\end{equation}

The only nonlinearity in this system is in the unidirectional spring.
This is described in the spring's constitutive law (eq. \ref{eq:spring2}).
In the interest of clarity we introduce the term $\Delta y = y_p - y_r$, to represent the deflection of the spring.
Similarly, $\Delta\dot{ y}$ is defined as $\Delta\dot{ y} = \dot{y_p} - \dot{y_r}$.
The damping is single acting because a majority of the damping is in the spring itself.
To make the force continuous and differentiable with respect to position a middle case is added.
Here $\epsilon$ is a small deflection.
\begin{equation}
F_{sp} = \begin{cases} 
      k\Delta y + b\Delta\dot{ y} & \Delta y > 0\\
      k\Delta y \hspace{-1 pt}-\hspace{-2 pt} b(1 \hspace{-1 pt} - \hspace{-2 pt} 2(\frac{\Delta y}{\epsilon})^3 \hspace{-1 pt} -\hspace{-2 pt}  3(\frac{\Delta y}{\epsilon})^2 )\Delta\dot{ y} &
      -\epsilon \leq \Delta y \leq 0\\
      0 & \Delta y < -\epsilon
   \end{cases}
    \label{eq:spring2}
\end{equation}

The values of the model parameters are tabulated below. 
$b$ was calculated from an observed damping ratio of 0.5 while the system oscillated under the weight of the robot.
\begin{center}
\begin{tabular}{ r | r l}
  Parameter & Value &\\   \hline
  $M$ & 11.07 & $kg$ \\
  $M_h$ & 0.45 & $kg$ \\  
  $g$ & 9.81 & $\nicefrac{m}{sec^2}$ \\
 $k$ & 5250 & $\nicefrac{N}{m}$ \\
 $b$ & 300 & $\nicefrac{Nsec}{m}$\ \\
\end{tabular}
\end{center}

\section{Controller Design}
\label{sec:ControllerDesign}
The desired behavior of this system is separated into three cases, each with its own controller. 

The first is transparent operation. In this case, the system will minimize the disturbance to the robot. Additionally, the system will stay near the robot in case intervention is necessary. 
In the second case, we actively control the supportive force on the robot. In the final case, when failure is sensed, the system will catch and lift the robot in order to prevent damage.

\subsection{Shadowing Controller}
The shadowing controller's purpose is to stay close to the robot in case it fails, while not affecting the dynamics of robot. 
To accomplish this, the planarizer will be commanded to always stay a small distance below the robot.
A proportional controller with velocity feed forward is used (eq. \ref{eq:PD2}).
The desired position is a small distance below the current robot position, while the desired velocity is the current robot velocity.
$\dot{y}_{motor}$ is the commanded planarizer velocity and $d$ is the desired offset between the robot and planarizer. 
\begin{equation}
\dot{y}_{motor} = k_p(y_r - d - y_p) + k_{\text{ff}}\dot{y}_r
    \label{eq:PD2}
\end{equation}
The feed forward term on the robot velocity was included to improve the tracking performance, which is critical to maintaining a constant offset between the robot and the planarizer.
To prevent overshoot, no integral gain was included in the controller.

\subsection{Force Controller}
The force controller applies a vertical desired force to the robot.
We use the measured force from the load cell for feedback control.
This measurement is noisy so we applied a first order low pass filter with a cutoff frequency of $\omega_c$ (eq. \ref{eq:ForceFilter}).
\begin{equation}
\frac{F(s)}{F_{in}(s)} = \frac{1}{s/\omega_c + 1}
    \label{eq:ForceFilter}
\end{equation}
The force is controlled using a proportional controller on filtered force while feeding forward the robot velocity (eq. \ref{eq:ForcePD}).
Here $F$ is the filtered force applied to the robot and $F_{des}$ is the desired force.
\begin{equation}
    \dot{y}_{motor} = k_p(F_{des} - F) + k_{\text{ff}}\dot{y}_{r}
    \label{eq:ForcePD}
\end{equation}
The feed forward term reduces the effect of robot movement.
We chose to use only proportional control and avoid derivative control because of the noise in the load cell measurements.

\subsection{Failure Detection}
To avoid damage to the robot, the recovery controller must be triggered automatically.
If the triggering criteria is too conservative, it will falsely trigger in situations where the robot can still operate.
This could prevent the robot from being able to access its full viable state space.
Alternatively, if the triggering criteria is too lenient it could trigger too late and risk damaging the robot.

\begin{figure} 
    \includegraphics[width=3.25in]{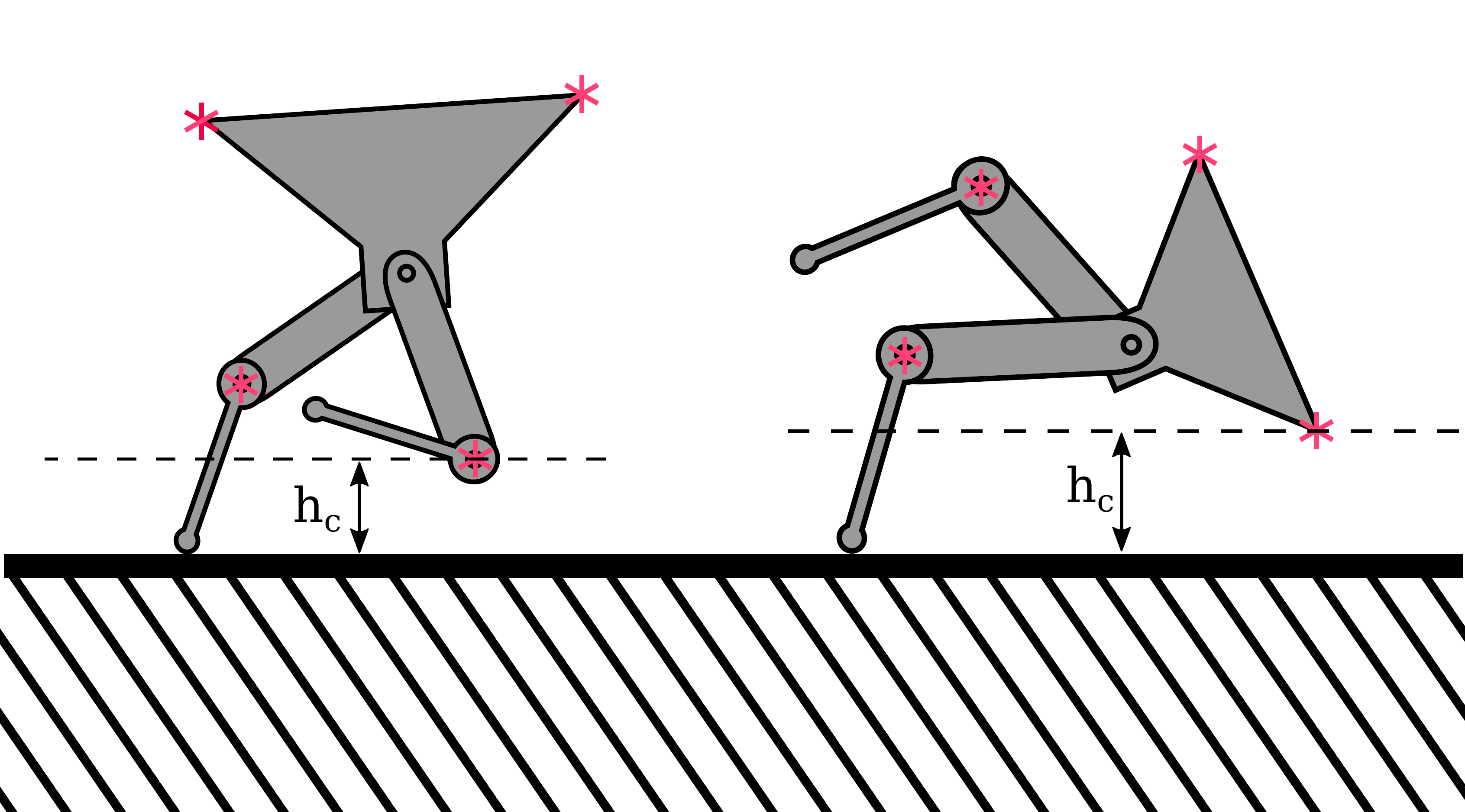}
    \caption{The undesirable ground contact points of RAM\emph{one}. These points are the knees and the top corners of the main body. The lowest of these points is checked against a minimum acceptable height to determine if recovery is necessary.}
    \label{fig:FailureConditions}
\end{figure}

The first definition of failure is based on unwanted ground contact.
Ground contact from any body part except the feet should be avoided.
This condition is reduced to checking if several points on the robot are too close to the ground.
In the case of RAM\emph{one}, there are only four points that need to be checked.
We check each of the knees and the farthest out corners of the main body. 
Failure is triggered if the lowest of these points is below the minimum allowable height (fig. \ref{fig:FailureConditions}).
For robots with different topologies or limb shapes, the undesired ground contact conditions must be reassessed.

We observe that when RAM\emph{one} has fallen in the past, it normally enters a fault mode before it gets close to the ground.
When a joint angle leaves the safe range, a system fault is triggered.
This halts all robot motors to ensure the limbs do not damage themselves by driving into a hard stop.
The system uses this as a second definition of failure.

When either type of failure is detected, first the robot motors are halted.
Second the system is switched from the shadowing or force controller to the recovery controller.

\subsection{Recovery Controller}
Once failure has occurred, we need to lift the robot clear of danger. 
To do this, we need to control the robot's position, so the dynamics of the system will have to be considered.

When lifting the robot, we would like to always have the ability to apply a force to the robot.
This is to avoid the case in which the robot is moving upward very fast and cannot stop before colliding with hardware at the top of the planarizer's travel.
When we are lifting the robot we can assume that the legs are off the ground and that $F_{ext} = 0$. 
Any force we apply to the robot passes through the spring, which means that the ability to apply a force to the robot is based on the current spring deflection.
We restrict our system to only operate in the linear region $\Delta y > 0$ or, $ y_c > y_p $.

Applying this condition to the equation of motion, the resulting requirement is a kinematic one.
The robot acceleration must be restricted according to (eq. \ref{eq:linCond}). Otherwise, it will lift off of the spring.
\begin{equation}
\ddot{y}_r > -g\frac{M-M_h}{M} + \frac{B}{M}
    \label{eq:linCond}
\end{equation}
$$B \geq |b\Delta\dot{ y}|$$

In this equation $B$ is an upper bound on the damping force. 
The damping was treated as a disturbance and bounded to allow the final kinematic requirement to be independent of velocity.
We can see that we will stay in our linear region as long as we keep our robot acceleration below a given constant.
To obtain an estimation of $B$, we estimate that $\Delta\dot{ y} < 0.1$ m/s during operation.
This estimate was confirmed by analysis of $\Delta\dot{ y}$ after implementation.
This yields $B = 30$ N, and a kinematic condition of $\ddot{y}_r > -6.7$ m/sec$^2$.
This is the condition that must be guaranteed during operation of the recovery controller.

We need to impose this acceleration limit on the motion of the system.
Our approach was to separate this into a low level controller, and a trajectory generator. 
We can create a trajectory for the catch system that satisfies the acceleration constraints that we are imposing.
The low level controller can then be tuned so that it accurately tracks the trajectory without concern for the kinematic constraints.

\begin{figure} 
    \includegraphics[width=\columnwidth]{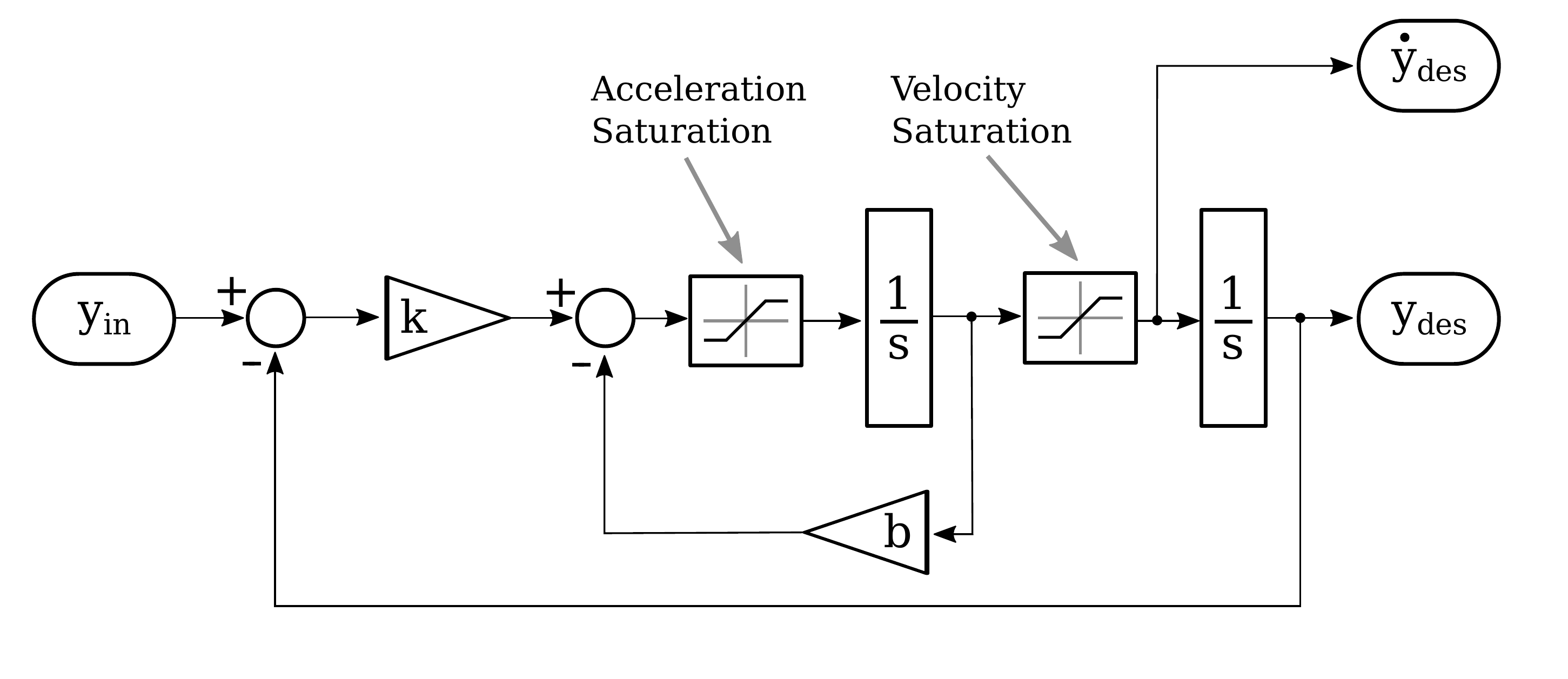}
    \caption{The set point filter block diagram. This creates a trajectory with a smooth, acceleration limited, velocity limited profile. The output is a desired position and velocity.}
    \label{fig:setPointFilter}
\end{figure}

The smooth trajectory is generated by a nonlinear set point filter. This is formed by taking a linear set point filter (eq. \ref{eq:setPtFilter}) and adding saturation (fig. \ref{fig:setPointFilter}).
\begin{equation}
\frac{Y_{des}(s)}{Y_{in}(s)} = \frac{1}{(\tau s + 1)^2} = \frac{k}{s^2 + bs + k}
    \label{eq:setPtFilter}
\end{equation}
Where $k = \frac{1}{\tau^2}$ and $b = \frac{2}{\tau}$. 
The linear version generates a smooth trajectory, but it does not limit acceleration.
The signal is saturated before the integrations occurs (fig. \ref{fig:setPointFilter}).
This limits maximum acceleration and velocity.
The previous analysis showed that a limit on velocity is not required, but it is beneficial for safety to explicitly limit the maximum speed that the system will move.

When the recovery mode is first enabled, the set point filter needs to be initialized.
The filter's integration blocks are initialized with the current position and current velocity of the robot.
This forces the generated trajectory to start with the current robot position and velocity.

For the low level controller, proportional control with feed forward is used (eq. \ref{eq:PD1}).
The set point filter provides both $y_{des}$ and $\dot{y}_{des}$ for use by this controller.
\begin{equation}
\dot{y}_{motor} = k_p(y_{des} - y_p) + k_{\text{ff}}\dot{y}_{des}
    \label{eq:PD1}
\end{equation}

This is a similar controller formulation to the shadowing controller.
However, the controller gains are not assumed to be the same.

\section{Results}
\label{sec:results}

\subsection{Shadowing Controller Results}
To test the behavior of the controller, the robot is manually moved up and down while the shadowing controller attempts to maintain an offset.
There are two metrics for evaluating the performance of the controller.
First, the total deviation in the offset should be less than the value of the offset to avoid loading the spring and applying significant forces to the robot.
Second, the acceleration of this offset should be small.
The offset corresponds to the height of the hanging mass, so when the offset accelerates the mass accelerates, thereby creating an undesirable force on the robot. 

The gains used for the shadowing controller are:
\begin{center}
\begin{tabular}{ c | r l}
  Parameter & Value &\\   \hline
  $d$ & 0.02 & m \\
  $k_p$ & 5 & sec$^{-1}$ \\  
  $k_{ff}$ & 1 & [ ] \\
\end{tabular}
\end{center}

\begin{figure} 
    \includegraphics[width=\columnwidth]{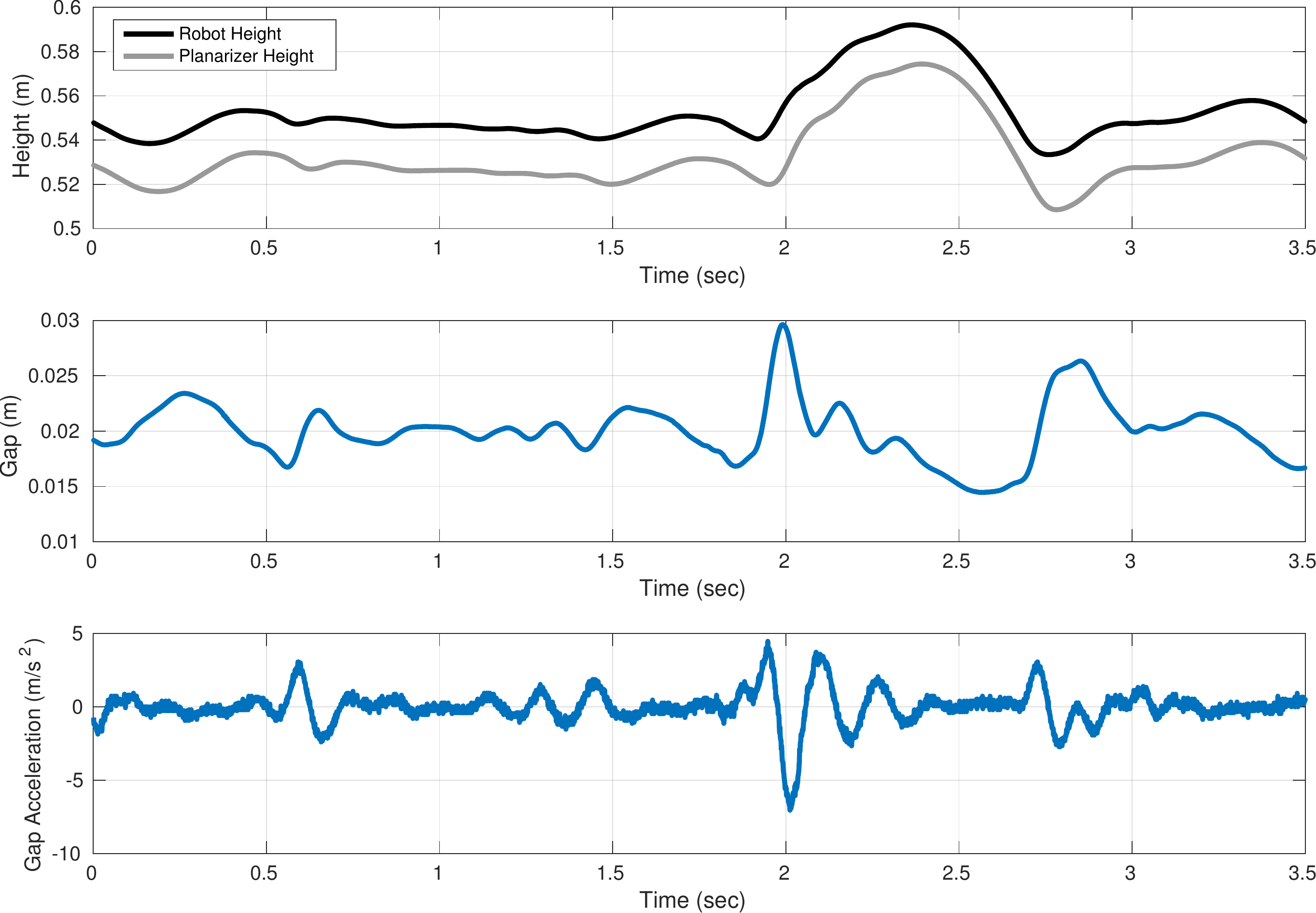}
    \caption{The behavior of the shadowing controller as the robot moves up and down.
    The overall gap deviation from the desired \unit[0.02]{m} is small enough that the spring is never loaded.
    The acceleration of the mass is small enough that the force created is less than 3\% of the robot weight.}
    \label{fig:following}
\end{figure}

The robot position and the planarizer position are shown (fig. \ref{fig:following}).
To see the tracking behavior more clearly, the gap between the two positions is shown over time.
The tracking error is never more than \unit[1]{cm}, which is less than our offset of \unit[2]{cm}.
In the final plot, the acceleration of the gap is shown over time.
from this we see that the magnitude of the acceleration is at most \unitfrac[7]{m}{s$^2$}.
This corresponds to a force of \unit[3.2]{N} on the robot, which is less than 3\% of the weight of the robot (an acceptably small disturbance).

\subsection{Force Controller Results}
\begin{figure} 
    \includegraphics[width=\columnwidth]{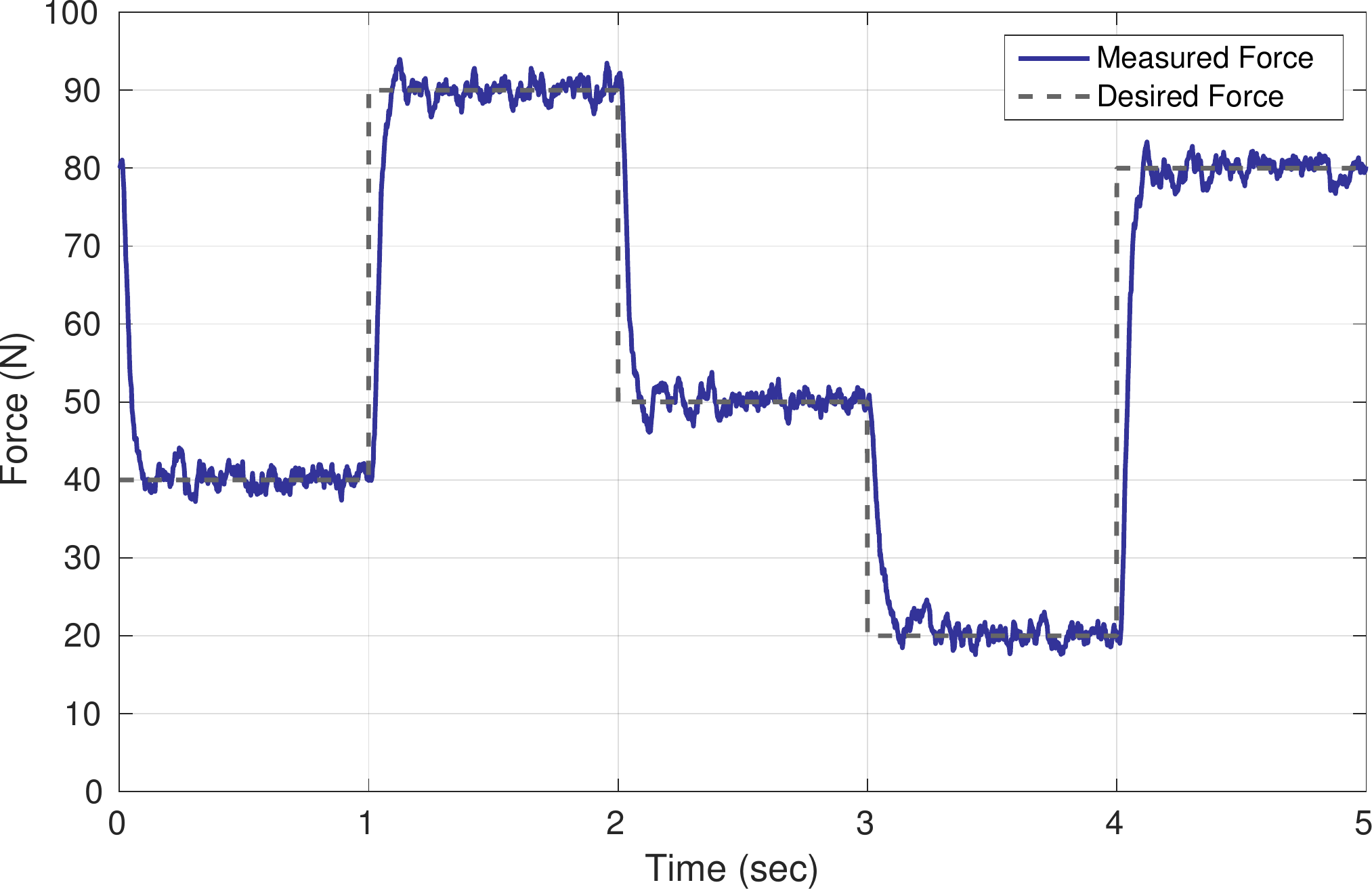}
    \caption{The behavior of the force controller as the desired force is varied.
    The measured force is reported after a low pass filter is applied.
    The persistent noise in the force signal prevents the controller from being tuned to be more aggressive.}
    \label{fig:forceResults}
\end{figure}

To test the force controller, a sequence of different desired forces were commanded for one second each.
During the test the robot was supported, so its height did not change significantly. 
The result of this sequence is shown (fig. \ref{fig:forceResults}).
This shows both the transient and steady state behavior of the controller.
The measured force given in the figure is the filtered signal (note that a significant noise is still present).
We elected to not lower the cutoff frequency because we did not want to further slow the response of the controller.
The gains used for the force controller are:
\begin{center}
\begin{tabular}{ c | r l}
  Parameter & Value &\\   \hline
  $\omega_c$ & 100 & $\nicefrac{rad}{sec}$ \\
  $k_p$ & 0.002 & $\nicefrac{m}{Nsec}$ \\
  $k_{ff}$ & 1 & [ ] \\
\end{tabular}
\end{center}

The largest jump in desired force resulted in a rise time of \unit[0.10]{sec}.
The overshoot is less than the variation in force measurements.
The variation in the steady state force is a result of noise not oscillation from the controller.
This noise corresponds to a force of \unit[$\pm$3.0]{N}.

\subsection{Recovery Controller Results}
We evaluated the recovery controller by its speed of intervention, and the presence of overshoot. 
The important timing metric is the delay between a failure command and when the robot is clear of danger.

The gains we used for the recovery controller are:
\begin{center}
\begin{tabular}{ c c | r l}
  Controller & Parameter & Value &\\   \hline
  Set Point Filter & $\tau$ & 0.0833 & sec \\ 
  &$v_{max}$ & 0.5 & m/sec \\
  &$a_{max}$ & 5 & m/sec$^{2}$ \\\hline
  P Controller &$k_p$ & 5 & sec$^{-1}$ \\
  &$k_{ff}$ & 1 & [ ] \\
\end{tabular}
\end{center}

\begin{figure} 
    \includegraphics[width=\columnwidth]{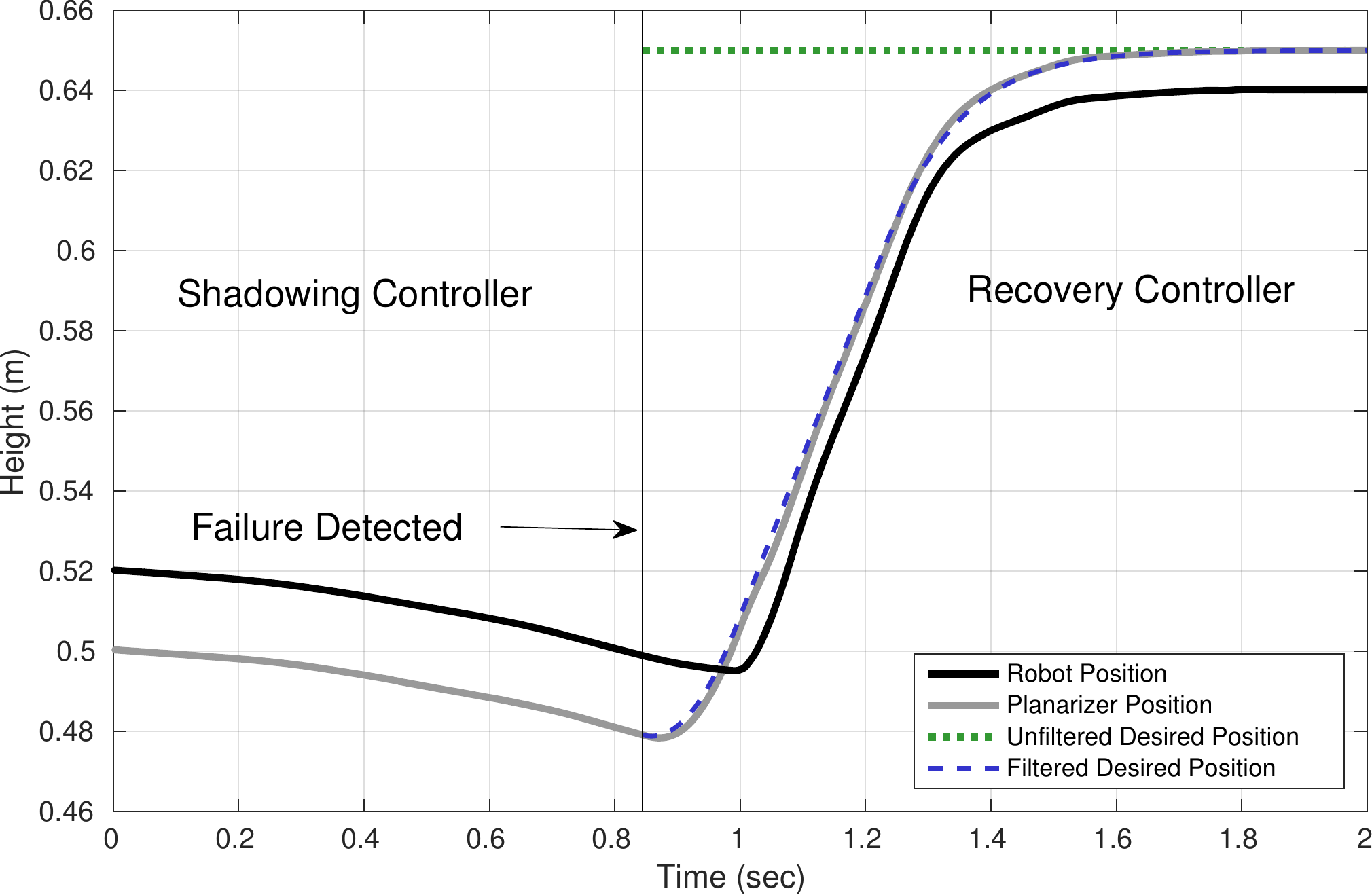}
    \caption{A failure event during a trial. 
    To the left of the thin vertical line, the shadowing controller is operating. 
    The thin black line signifies a failure event. 
    To the right of the line is the recovery controller bringing the robot up to a safe height.
    The trajectory generated by the set point filter is followed very accurately.}
    \label{fig:recoveryResult}
\end{figure}

Data from a failure event is shown (fig. \ref{fig:recoveryResult}). 
To the left of the thin vertical line is normal operation.
The shadowing controller is maintaining a \unit[2]{cm} offset.
The vertical line indicates where the failure event occurred.
In this case the failure arose from the knee moving too close to the ground.
The set point filter is initialized to the current planarizer position and velocity.
The smooth trajectory of the set point filter is followed very accurately.
There is a very small oscillation in the robot position as it approaches the safe height.
However, the robot does not overshoot its final position so this is acceptable.
In 0.4 seconds, the robot is \unit[8]{cm} above its position at failure, which is a safe height.

\section{Discussion \& Conclusion}
\label{sec:discussionandconclusion}
This paper presented the design and control of an active recovery system.
This system is able to provide safe and reliable failure recovery for a legged robot.
It can either minimally impact dynamics or apply an arbitrary force to a robot during normal operation.

Future work on this system include improved force control while the robot is moving.
A potential improvement to the force control algorithm could be the inclusion of spring deflection measurements as a second sensor.  
While this is only an indirect way of measuring force, it has potentially less noise than the load cell.

With some modification, this system can be applied to other planarizer configurations.
A commonly used planarizer in the literature is a rotating boom.
A winch can be mounted on the center pivot, a cable run up a central pole, through a pulley at the top and down to the robot.
A manual system similar to this was implemented for the planar robot Uniroo \cite{zeglin1991uniroo}.
Series elastic actuation can be added through a unidirectional spring.

An active recovery system can provide significant benefits to the development of legged robotic systems. 
Such a device can reduce the likelihood of damage to robots, increase trial speed, and apply assistive forces.
These benefits can be particularly useful for machine learning applications.
A force control policy, such as a height based potential field, could be used to provide support selectively during learning.
This system can allow continuous operation with minimal human intervention which is necessary for the large number of trials that machine learning algorithms require. 
This system allows legged systems to more efficiently explore their dynamics, with reduced concern for falls and collisions.

\bibliographystyle{ieeetr}
\bibliography{References}

\end{document}